\documentclass{article}
\usepackage{iclr2024_conference,times}

\usepackage{amsmath,amsfonts,bm}

\def\eqref#1{equation~\ref{#1}}

\def\1{\bm{1}}

\DeclareMathAlphabet{\mathsfit}{\encodingdefault}{\sfdefault}{m}{sl}
\SetMathAlphabet{\mathsfit}{bold}{\encodingdefault}{\sfdefault}{bx}{n}

\usepackage{hyperref}
\usepackage{url}

\usepackage{microtype}
\usepackage{graphicx}
\usepackage{xspace}
\usepackage{booktabs}
\usepackage{ulem}
\usepackage{multirow}
\usepackage{rotating}
\usepackage{algorithm}
\usepackage[noend]{algpseudocode}
\usepackage{subcaption}
\usepackage{paralist}
\PassOptionsToPackage{numbers, compress}{natbib}

\usepackage{amsmath}
\usepackage{amssymb}
\usepackage{mathtools}
\usepackage{amsthm}
\usepackage{hyperref}
\usepackage{wrapfig}

\usepackage[capitalize,noabbrev]{cleveref}

\theoremstyle{plain}

\theoremstyle{definition}

\theoremstyle{remark}

\usepackage[textsize=tiny]{todonotes}

\usepackage{subfiles} 

\def\method{Magnushammer\xspace}
\def\methods{\textsc{Select}\xspace}
\def\methode{\textsc{Rerank}\xspace}

\def\dsafpsh{\textsc{Machine-Augmented Proofs Library}\xspace}
\def\minif2f{miniF2F\xspace}

\makeatletter
\renewcommand\paragraph{\@startsection{paragraph}{4}{\z@}%
                       {-2\p@ \@plus -4\p@ \@minus -4\p@}%
                       {-0.5em \@plus -0.22em \@minus -0.1em}%
                       {\normalfont\normalsize\bfseries}}
\makeatother

\algnewcommand{\LeftComment}[1]{\Statex \(\triangleright\) #1}

\title{Magnushammer: A Transformer-Based \\Approach to Premise Selection}

\author{%
Maciej Mikuła\thanks{Equal contribution.}\hspace{1cm} \\ Google DeepMind\thanks{Work performed while at the University of Warsaw.}
\And
Szymon Tworkowski$^{*}$ \\ xAI$^{\dagger}$
\And
Szymon Antoniak$^{*}$\hspace{1cm} \\ Mistral AI$^{\dagger}$
\And
Bartosz Piotrowski \\ IDEAS NCBR \\ 
\And
Albert Qiaochu Jiang \\ University of Cambridge
\And
Jin Peng Zhou \\ Cornell University\thanks{Work performed while at Google Research.}
\And
Christian Szegedy \\ xAI$^{\ddagger}$
\And
Łukasz Kuciński \\ IDEAS NCBR
\And
Piotr Miłoś \\ IDEAS NCBR
\And
Yuhuai Wu \\ xAI$^{\ddagger}$
}

\iclrfinalcopy

\begin{document}

\maketitle

\begin{abstract}

   This paper presents a novel approach to premise selection, a crucial reasoning task in automated theorem proving. Traditionally, symbolic methods that rely on extensive domain knowledge and engineering effort are applied to this task. In contrast, this work demonstrates that contrastive training with the transformer architecture can achieve higher-quality retrieval of relevant premises, without the engineering overhead. Our method, \method, outperforms the most advanced and widely used automation tool in interactive theorem proving called Sledgehammer. On the PISA and \minif2f benchmarks \method achieves $59.5\%$ (against $38.3\%$) and $34.0\%$ (against $20.9\%$) success rates, respectively. By combining \method with a language-model-based automated theorem prover, we further improve the state-of-the-art proof success rate from $57.0\%$ to $71.0\%$ on the PISA benchmark using $4$x fewer parameters. Moreover, we develop and open source a novel dataset for premise selection,
   containing textual representations of (\textit{proof state}, \textit{relevant premise}) pairs. To the best of our knowledge, this is the largest available premise selection dataset, and the first one for the Isabelle proof assistant.
\end{abstract}


\section{Introduction}

\begin{wrapfigure}{r}{0.49\textwidth}

    \vspace{-24pt}
    \begin{center}
    \includegraphics[width=\linewidth]{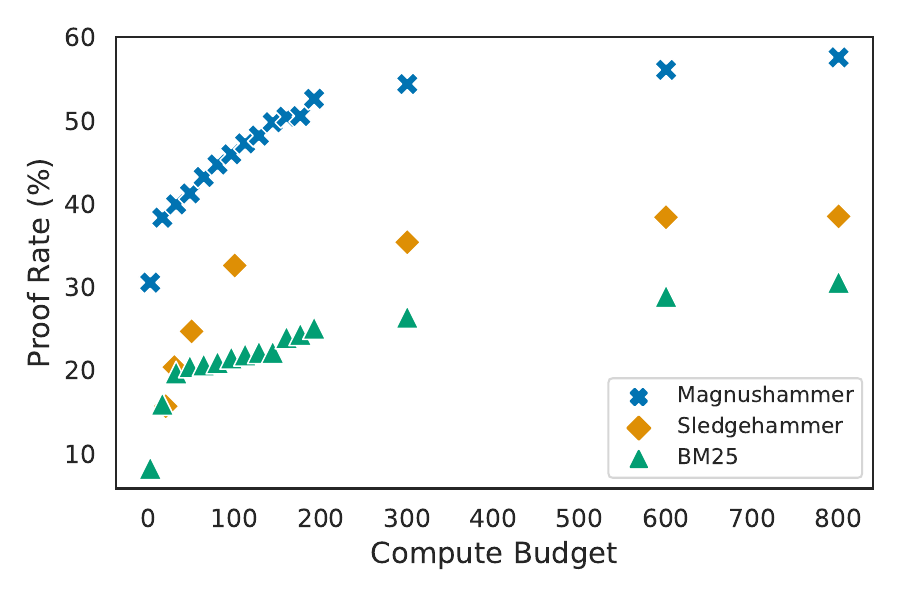}
    \end{center}
    \vskip -15pt
    \caption{
    Proof success rate for varying computational budget for \method, Sledgehammer, and BM25.
    \method shows remarkable scalability. See Sections \ref{sec:compute_budget_definition} for the definition of computational budget and Section \ref{sec:budget_experiments} for configurations depicted in this figure.}
    \label{fig:compute_cost_ablation}
    \vskip -0.2in
\end{wrapfigure}

Automating mathematical reasoning has been a central theme of artificial intelligence since its earliest days~\citep{de1970mathematical}.
Recently, machine learning has led to significant advancements in both informal ~\citep{lewkowycz2022solving} and formal mathematical reasoning~\citep{kaliszyk2015mizar,alemi2016deepmath, polu2020generative, han2021proof}.
The latter approach, adopted in this paper, allows mechanical verification of proofs by proof assistants.

Modern mathematics development is gradual:
it feeds upon a huge body of already established knowledge and constantly adds to it.
Proving a mathematical statement requires retrieval of facts from the knowledge base that can advance the proof. In automated reasoning literature, this retrieval process is known as \textit{premise selection}.

Many tools have been developed to tackle premise selection~\citep{alama2011premise, kuhlwein2012overview, kaliszyk2017holstep, bansal2019holist}, including a broad class known as ``hammers,'' which {leverage} powerful automated theorem provers (ATPs) to {determine useful premises}~\citep{Paulson2012ThreeYO, hol4hammer, holyhammer, coqhammer}. One such tool, Sledgehammer~(SH)~\citep{Paulson2012ThreeYO}, has gained prominence with Isabelle \citep{paulson2000isabelle}, where it helped to create a significant portion of Isabelle's proof corpus.
Hammers are not yet available in all proof assistants~\citep{ebner2020integration}: implementing them is challenging due to
the complex techniques required for different logics and type systems.
There is a need for an effective premise selection tool that requires less adaptation to work for different proof assistants.

In this study, we provide a generic, data-driven, transformer-based \citep{vaswani2017attention} premise selection tool: \method. It constitutes a novel way to tackle the premise selection task, effective while requiring little domain-specific knowledge. \method is trained contrastively to perform premise retrieval in two stages: in the \methods stage, it retrieves the most relevant $1024$ premises (measured by the cosine similarity of their embeddings to that of the current proof state) from tens of thousands {(the database contains 433K premises in total and typically 30K--50K are available in each proof state)};
in the \methode stage, the retrieved premises are re-ranked with proof-state-aware scores:
tokens of the proof state directly attend to tokens of the premise, giving a more contextualized relevance score. An overview of \method's architecture is shown in Figure~\ref{fig:select_expand_overview}.

\method can prove $59.5\%$ of the theorems on the PISA benchmark \citep{jiang2021lisa}, a substantial improvement over Sledgehammer’s $38.3\%$. We demonstrate that this dominance is consistent with varying controlled compute budgets, shown in Figure~\ref{fig:compute_cost_ablation}.
Furthermore, we replace the premise selection component~(Sledgehammer) in a neural-symbolic model Thor~\citep{jiang2022thor} with \method and improve the state-of-the-art proof success rate on PISA from $57\%$ to $71\%$.

To train \method, we extracted a premise selection dataset from the Isabelle theorem prover and its human proof libraries.
The dataset consists of $4.4$M premise selection instances, with $433$K unique premises.
To the best of our knowledge, this is the largest open-sourced premise selection dataset, and the first one of this kind for Isabelle. We find \method to be data efficient, outperforming Sledgehammer with only $4$K training examples ($0.1\%$ of the training data available).

\paragraph{Contributions}
\begin{itemize}

    \item We propose the use of transformers trained contrastively as a novel way of addressing the premise selection problem. Our method, \method, achieves a $59.5\%$ proof rate on the PISA benchmark, significantly improving the $38.3\%$ proof rate of Sledgehammer,
    the most powerful general-purpose automation tool for Isabelle.
    \item We extract and open source the largest, to the best of our knowledge, premise selection dataset. It consists of $4.4$M premise selection examples and $433$K unique premises.

    \item We analyze how \method's performance depends on the model size, dataset size, and the inference-time compute budget. We show its superiority with moderate resources.


\end{itemize}


\section{Background: proof assistants, Isabelle, and Sledgehammer}

\label{sec:sledgehammer}
Proof assistants (aka interactive theorem provers, or ITPs)
such as
Isabelle \citep{paulson2000isabelle},
Lean \citep{moura2015lean},
Coq \citep{coq},
HOL Light \citep{hollight},
or Mizar \citep{mizar},
are software tools designed to assist the development of formal proofs.
They provide expressive language for the formalization of mathematical statements and proofs while verifying them formally.
In Isabelle, theorems are proved sequentially:
an initial \textit{proof state} is obtained after the theorem is stated, and the proof state changes when
the user provides a valid \textit{proof step} (see Appendix \ref{app:visualization_of_isa_env} for an example).
Proof states contain information about the already established facts and the remaining goals to prove. Proof steps consist of \textit{tactics}, which are optionally parametrized by \textit{premises}.
Tactics are theorem-proving procedures and can complete some proofs in one step
provided with relevant premises. However, finding these premises is difficult:
one needs to select a handful of relevant facts from the current proof context,
which typically contains tens of thousands of them.


\begin{figure}[t]
    \centering
    \begin{subfigure}{.47\textwidth}
    \includegraphics[width=\linewidth,trim=4 4 4 4,clip]{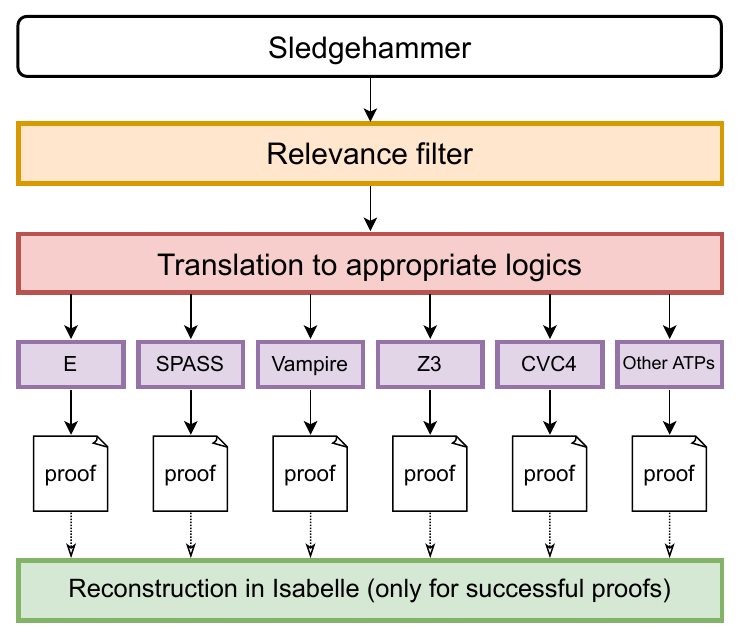}
    \caption{\small A call to Sledgehammer triggers the following sequence of steps: First, available facts are filtered based on their similarity to the conjecture. Then, the conjecture together with the selected facts (usually a few hundred in number) are translated to simpler logic used by the external provers (E, SPASS, etc.). Then, such problems are fed into each ATP separately.
    Finally, the premises used in the successful ATP proofs are used
    to reconstruct a proof inside Isabelle using its native methods.
    }
    \label{fig:secSH}
    \end{subfigure}
    \hfill
    \begin{subfigure}{.47\textwidth}
    \centering
    \includegraphics[width=\linewidth,trim=4 4 4 4,clip]{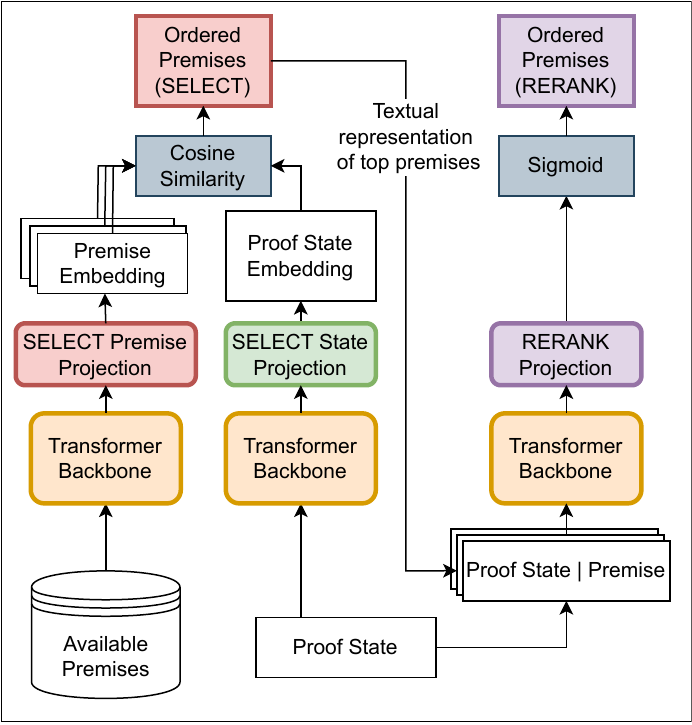}
    \caption{\small Given a proof state, we first retrieve the most relevant premises according to the cosine similarity of their embeddings with the proof state embedding~(\methods). We then re-rank these with a model that encodes each proof state and premise pair, outputting a relevance score (\methode). The bulk of the architecture is a shared transformer model, in orange.}
    \label{fig:select_expand_overview}
    \end{subfigure}

    \caption{
    Overview of Sledgehammer (a) and \method (b).
    }
\end{figure}

Sledgehammer~\citep{Paulson2012ThreeYO, smt_solvers_blanchette} is a powerful automated reasoning tool for Isabelle. It belongs to a broader class of tools known as ``hammers,'' which integrate automated theorem provers (ATPs) into proof assistants. The goal of these tools is to support the process of finding and applying proof methods. Sledgehammer has become an indispensable tool for Isabelle practitioners~\citep{Paulson2012ThreeYO}. It allows for closing low-level gaps between subsequent high-level steps of proof without the need to memorize entire lemma libraries. 

Sledgehammer is designed to first pre-select a number of relevant facts heuristically, translate them together with a conjecture to simpler logic, and try to prove the conjecture using strong, external ATPs like E~\citep{48}, SPASS~\citep{56}, Vampire~\citep{44},
Z3~\citep{33}, or cvc5~\citep{cvc5}. If successful, these provers generate complete proofs. They are, however, not trusted by Isabelle. Instead, the facts used in the external proofs are extracted and used to produce a proof \textit{inside} Isabelle using its native methods. Up to this last step, known as \textit{proof reconstruction}, Sledgehammer is essentially used as a precise premise selection tool.
See Figure \ref{fig:secSH} depicting the whole process.


While immensely useful, Sledgehammer comes with several limitations. First, increasing computational power for Sledgehammer brings quickly diminishing returns~\citep{judgement_day}.
Second, the logic projection and proof reconstruction in a hammer are not straightforward for type systems other than higher-order logic~\citep{coqhammer}. Finally, Sledgehammer's performance hinges on the relevance filtering scheme, a suite of methods based on handcrafted heuristics~\citep{meng2009lightweight} or classical machine learning~\citep{kuhlwein2013mash}. Such approaches are unlikely to efficiently utilize the constantly growing body of proof data.

We argue that all these limitations can be overcome with deep-learning-based approaches. Neural networks have shown remarkable effectiveness in end-to-end problem solving with little or no feature engineering~\citep{krizhevsky2012imagenet, gpt3}. Adopting textual representations with generic neural solutions removes the need for logic projection, ATP solving, and proof reconstruction.
Moreover, large language models have recently displayed impressive scaling properties with respect to both model size~\citep{kaplan2020scaling} and data~\citep{chinchilla}.

\section{\method} \label{sec:our_method}

The goal of premise selection is to find relevant mathematical facts for a given proof state. We focus on selecting premises with a neural model informed by their textual representations instead of relying on fact structures like Sledgehammer~(see Section \ref{sec:sledgehammer}). The core idea of \method is to combine fast retrieval based on representational similarity (\methods) with a more accurate re-ranking (\methode), as outlined in Algorithm \ref{alg:selectAndExpand}. Our method closely follows those of~\citet{nogueira2019passage} and~\citet{contriver}. This hierarchical approach is scalable to large formal libraries containing hundreds of thousands of facts. Below we describe the two-stage \method approach.

\methods leverages \textit{representation similarity} and is based on batch-contrastive learning similar to the methods of \citet{alemi2016deepmath}, \citet{bansal2019holist}, \citet{han_informal_premise_selection}, or \citet{radford2021learning}. \textsc{Select} embeds premises and proof states into a common latent space and uses cosine similarity to determine their relevance. During inference, it requires only one pass of a neural network to compute the proof state embedding and dot product with cached premise embeddings. \textsc{Select} is hence fast and scalable to large sets of premises. In our experiments, there are between $30$K and $50$K premises in a typical proof state context, from which we select $K_{S} = 1024$ most relevant ones.

\methode scores the relevance of the $K_{S}$ selected premises for the current proof state by analyzing the $(\mathtt{proof\_state}, \mathtt{premise})$ pairs.
\methode is trained to output the probability of the $\mathtt{premise}$ being relevant to the $\mathtt{proof\_state}$. The $K_S$ premises retrieved by \methods are re-ranked with respect to these probabilities, and the final list comprises of the top $K_{R}$ premises (we set $K_R = K_S$). Having both the premise and the proof state in a single input allows \methode to be more accurate. However, at the same time, it is much slower, as each pair must be scored individually.

\begin{algorithm}[H]
   \caption{Premise selection with \method.}
   \label{alg:selectAndExpand}
\begin{small}
\begin{algorithmic}[1]
\Require
\Statex $\mathtt{proof\_state}, \mathtt{premises}$ \Comment{proof state to retrieve premises for and database of available premises}
\Statex $K_{S}, K_{R}$
\Comment{number of premises to retrieve with \methods and \methode, respectively}
%
\State $\mathtt{state\_embedding} \gets  \mathtt{get\_embeddings}(\mathtt{proof\_state})$
\Comment{\methods stage starts}
\State $\mathtt{premises\_embeddings} \gets  \mathtt{get\_embeddings}(\mathtt{premises})$
\State $\mathtt{Cache}(\mathtt{premises\_embeddings})$
\State $\mathtt{sim\_scores} = \mathtt{state\_embedding} \cdot \mathtt{premises\_embeddings}$
\State $\mathtt{selected} = \mathtt{premises}[\mathtt{argsort(-sim\_scores)}[:K_{S}]]$
\State $\mathtt{batch} = []$
\Comment{\methode stage starts}
\For{$\mathtt{premise}$ in $\mathtt{selected}$}
\State $\mathtt{batch.append}((\mathtt{proof\_state}, \mathtt{premise}))$
\EndFor
\State $\mathtt{rerank\_scores} \gets \mathtt{get\_rerank\_scores}(\mathtt{batch})$
\State $\mathtt{top\_premises} = \mathtt{selected}[\mathtt{argsort(-rerank\_scores)}[:K_{R}]]$
\State \textbf{return} $\mathtt{top\_premises}$
\end{algorithmic}
\end{small}
\end{algorithm}

\paragraph{Training} We train \method using two alternating tasks: \methods is trained with a modified InfoNCE loss \citep{oord2018representation}, and \methode is trained with the standard binary cross-entropy loss.
The architecture of \method shares a transformer backbone with specialized linear projections on top (see Figure \ref{fig:select_expand_overview}). The backbone is pre-trained with a language modeling task on the GitHub and arXiv subsets of the Pile dataset \citep{gao2021pile}. For training, we use datasets consisting of $(\mathtt{proof\_state}, \mathtt{premise})$ pairs extracted with a procedure described in Section \ref{sec:datasets}.

During \methods's training, each batch consists of $N$ proof states, $N$ positive premises (one for each proof state), and additional $M$ negative premises sampled from available facts that are not ground truth premises for any of the selected proof states. This gives $N - 1 + M$ negatives per proof state in one batch. We typically use $M=3N$, which differs from standard batch-contrastive learning \citep{radford2021learning}, in which $M = 0$ and negatives are only the other $N-1$ premises in the batch
\methode is trained using a binary classification objective. For each positive $(\mathtt{proof\_state}, \mathtt{premise})$ pair in the dataset, we construct $15$ negatives from the most likely false positives returned by \methods. Specifically, all the premises $\mathcal{M}$ that are facts that were never used as a premise for $\mathtt{proof\_state}$, are first chosen. Then, the top $1024$ of $\mathcal{M}$ according to \methods are selected, and $15$ are sampled from them to construct negative training pairs. See Appendix \ref{sec:selectAndExpandAppendix} for complete training details.

\paragraph{Evaluation in Isabelle}\label{sec:psm_eval_in_isabelle}
We outline how premises chosen by \method are used to prove theorems in
Isabelle. Given a proof state, a list of the $k$ most relevant
premises $P$ is retrieved. We construct proof steps consisting of a tactic $t$ and a subset
of premises $S\subseteq P$. Such proof steps are executed in parallel, with a
timeout of $2$ seconds. The evaluation is successful if any of these proof
steps completes the proof. For $S$, we pick the top $i$ of $P$, where $i$'s are
consecutive powers of $2$ up to $2^{10}$, or $0$ for tactics that do not accept
premises.
More details, including the set of tactics used, are presented in Appendix~\ref{app:eval_on_theorems}. An example of a proof with tactics and premises is given in Appendix~\ref{app:proof_with_premises}.

Note that the procedure of trying multiple different subsets of premises is commonly applied in the context of automated theorem proving \citep{malaria2008,kuhlwein2012overview} and similar to the technique implemented in Sledgehammer~\citep{Paulson2012ThreeYO}. The rationale behind this is that the proof procedures implemented in ATPs and high-level ITPs' tactics perform combinatorial search, and providing them with fewer premises to restrict their search space is beneficial.

\section{Datasets} \label{sec:datasets}

We created and released\footnote{\scriptsize\url{https://huggingface.co/datasets/Simontwice/premise_selection_in_isabelle}} a comprehensive dataset of textual representations for Isabelle's proof states and premises.
To the best of our knowledge, this is the first high-quality dataset of this kind for Isabelle, and also the largest premise selection dataset overall. We used the two largest collections of Isabelle theories to create the dataset: the \href{https://www.isa-afp.org/}{Archive of Formal Proofs} and the \href{https://isabelle.in.tum.de/library/}{Isabelle Standard library}.


For every proof step in every proof from these collections, we extracted the preceding proof state and the set of premises used in the proof step; this was turned into  $(\mathtt{proof\_state, premise})$ pairs constituting training data points. We call this the \textsc{Human Proofs Library} (HPL) dataset. In addition, we used Sledgehammer to generate proofs that are different from the human ones by using potentially alternative premises. We refer to this as the \textsc{SH} partition, and its union with HPL constitutes the \dsafpsh~(\textsc{MAPL}) dataset. Statistics for all these datasets  are given in Table~\ref{tab:dataset_ps}. Note that MAPL grosses over $4$M data points.

\begin{wraptable}{r}{0.47\textwidth}
  \small
  \centering
  \setlength{\tabcolsep}{0.7em}
    \caption{Statistics of MAPL and both its partitions: HPL (coming from human-written proofs) and SH (coming from Sledgehammer-generated proofs). The data points are of the form of $\mathtt{(proof\_state, premise)}$ pairs.
    }
  \label{tab:dataset_ps}
  \begin{tabular}{lcccc}
  \toprule
  Dataset & \textsc{HPL} & \textsc{SH} & \textsc{MAPL} \\
  \midrule
  Data points & 1.1M & 3.3M & 4.4M\\
  Unique proof states & 570K & 500K & 570K\\
  Unique premises & 300K & 306K& 433K \\
  \bottomrule
  \end{tabular}
\end{wraptable}

Below we describe in more detail how data points are extracted from a proof step.
An Isabelle's proof is a sequence of
$(\mathtt{proof\_state}, \mathtt{proof\_step})$ pairs: $\mathtt{proof\_state}$ has the state information, and $\mathtt{proof\_step}$ is a tactic application that advances the proof. A $\mathtt{proof\_step}$ may use $\mathtt{premises}$: theorems, lemmas, or definitions established previously. Suppose a $\mathtt{proof\_step}$ contains $n$ premises: $p_1, p_2, \ldots, p_n$. We then extract $n$ data points: $(\mathtt{proof\_state}, p_1), \ldots, (\mathtt{proof\_state}, p_n)$.
Executing Sledgehammer on the $\mathtt{proof\_state}$ may result in multiple different synthetic $\mathtt{proof\_step}$s, and data points can be extracted from each in the same way (see Appendix \ref{app:alternative_proof_step_generation_with_sh} for details).

Mining the HPL partition took $10$K CPU hours, and mining the SH partition took $150$K CPU hours (17 CPU years) on a distributed system.

Our datasets have $2$ distinguishing features:
\begin{compactenum}
    \item The human-originating dataset is augmented by alternatives generated with  Sledgehammer, which results in a significantly larger and more diverse dataset.
    This also decreases the probability of sampling \textit{false negatives} while training contrastively: a negative example $(\mathtt{proof\_state}, \mathtt{premise})$ may in fact be positive, but we just have not seen an alternative proof using $\mathtt{premise}$. Generating multiple alternative proofs partially remedies this problem.
    \item Both $\mathtt{proof\_state}$s and $\mathtt{premise}$s are represented as
    ``high-level'' Isabelle's text
    instead of ``low-level'' logical formalism like, e.g., TPTP~\citep{tptp} used by \citet{alama}.
    This makes the dataset more suitable for language models, decreases the
    need for feature engineering, and facilitates cross-proof-assistant
    pre-training~\citep{conneau2019cross}.
\end{compactenum}

\section{Experiments} \label{sec:experiments}

We evaluate \method on the PISA and \minif2f theorem proving benchmarks using  \textit{proof success rate} as a metric. Our main result is that \method{} outperforms Sledgehammer by a large margin and, combined with Thor \citep{jiang2022thor}, sets a new state of the art on the PISA benchmark ($71.0\%$ from $57.0\%$). Through ablations, we study the effectiveness of \method and the contribution of its components. Additional results and details can be found in Appendix \ref{app:additionalResults}.

\subsection{Experimental details}\label{sec:exp_details}

\paragraph*{Benchmarks}
For evaluation, we use PISA~\citep{jiang2021lisa} and \minif2f~\citep{zheng2021minif2f} benchmarks.
PISA contains problems randomly selected from the Archive of Formal Proofs;\footnote{When training on data from the Archive of Formal Proofs, we remove the subset of it appearing in PISA.} we use the same $1000$ problems as \citet{jiang2022thor} for our evaluations. \minif2f consists of $488$ high-school competition-level problems, split into validation and test set, each with $244$ problems.

\begin{table}
    \centering
    \caption{Proof rates on the PISA benchmark. On the single-step task,
    \method outperforms both Sledgehammer and BM25 by a wide margin. On
    the multi-step task, \method combined with Thor achieves the
    state-of-the-art proof rate of $71.0\%$.}

    \begin{tabular}{llc}
    \toprule
    Task & Method   & Proof rate (\%) \\
    \midrule
    & BM25 & $30.6$ \\
    & TF-IDF & $31.8$ \\
    Single-step&OpenAI embed. \small~\citep{neelakantan2022text}& $36.1$ \\
    &Sledgehammer & $38.3$ \\
    &\method (ours) & $\textbf{59.5}$ \\
    \midrule
    &LISA\textnormal{\small~\citep{jiang2021lisa}} & $33.2$ \\
    Multi-step &Thor\textnormal{\small~\citep{jiang2022thor}} & $57.0$ \\
    &Thor + \method (ours) & $\textbf{71.0}$ \\
    \bottomrule
    \end{tabular}
    \label{tab:main_result}
\end{table}


\begin{table}
    \centering
    \caption{Proof rates on the \minif2f benchmark. On the single-step task, \method outperforms Sledgehammer and its variant with additional heuristics \citep{jiang2022draft}.  On the multi-step task, Thor+\method obtains competitive results, significantly outperforming Thor+Sledgehammer.}
    \label{tab:minif2f_results}
    \begin{tabular}{llcc}
    \toprule
    Task&Method & Valid~(\%) & Test~(\%)\\
    \midrule
    &Sledgehammer & $9.9$ & $10.4$ \\
    Single-step&Sledgehammer + heuristics & $18.0$ & $20.9$~\\
    &\method{} {{(ours)}} & $\textbf{33.6}$ & $\textbf{34.0}$ \\
    \midrule
    &Thor + Sledgehammer\textnormal{\small~\citep{jiang2022thor}} & $28.3$ & $29.9$ \\
    Multi-step&Thor + Sledgehammer + auto \textnormal{\small~\citep{autoform_wu}} & $37.3$ & $35.2$ \\
    &Thor + \method{} (ours) & $36.9$ & $37.3$ \\
    &DSP\textnormal{\small~\citep{jiang2022draft}} & $\textbf{43.9}$ & $\textbf{39.3}$ \\
    \bottomrule
    \end{tabular}
\end{table}

%

\paragraph*{Metric and evaluation setups}
To evaluate the performance, we measure \textit{proof success rate}: the percentage of successful proofs. A proof is successful if it is formally verified by Isabelle. We distinguish \textit{single-step} and \textit{multi-step} settings. In the single-step setting, we check if the theorem can be proven in one step by applying premises retrieved by the evaluated premise selection method (e.g., \method{}).  In the multi-step scenario, we perform a proof search using a language model following Thor \citep{jiang2022thor}.
Thor + \method{} uses \method{} instead of Sledgehammer as the premise selection component. A further explanation is given in Section \ref{sec:main_result_multi}.

\paragraph{Evaluation protocol and computational budget}\label{sec:compute_budget_definition} 

Algorithm \ref{alg:evaluate_psm_on_proof_rate_benchmark} (in Appendix \ref{app:eval_on_theorems}) details the evaluation of \method{} in the single-step setting. It generates $|\mathcal{T}|\times |K|$ proof steps by combining each tactic $t \in \mathcal{T}$ with top $k$ premises from a ranking provided by \method{}, where $\mathcal{T}$ is a prescribed set of tactics, $k \in K$, and $K$ is a list of integers. Such constructed proof steps are then executed in Isabelle. We define the computational budget for such an evaluation as $C = |\mathcal{T}|\times |K|\times T$, where $T$ is a timeout expressed in seconds (we use $T=2$ s as we observed little benefit from increasing it).
Estimating the computational budget for Sledgehammer is difficult due to its complex internal architecture. We approximate it by $C=S\times T$, where $S$ is the `number of CPU cores' (corresponding to steps executed in parallel) and $T$ is the timeout. We use $S=10$ for our calculations. See Appendix~\ref{app:sledgehammer} for more details.

\paragraph*{Architecture and training details}
For our main experiments, we pre-train standard decoder-only transformer models with $38$M and $86$M non-embedding parameters and fine-tune them for downstream tasks of premise selection or proof step generation. Full details are given in Appendix \ref{app:trainingDetails}.
In our experiments, we use the Portal-to-ISAbelle API \citep{jiang2021lisa} to interact with Isabelle. 

\subsection{Results on PISA and miniF2F benchmarks}\label{sec:main_result_single}

Our main empirical results, summarized in Table \ref{tab:main_result} and Table \ref{tab:minif2f_results}, were obtained with the $86$M parameter model. Figure~\ref{fig:compute_cost_ablation} and Section \ref{sec:budget_experiments} deepen this study, showing that \method{} outperforms Sledgehammer across a broad spectrum of computational budgets.  

\paragraph{Performance on the single-step task}
In the single-step setting, \method outperforms Sledgehammer by a wide margin on both PISA ($59.5\%$ vs. $38.3\%$) and \minif2f ($34.0\%$ vs. $20.9\%$).
Additionally, on PISA, \method outperforms TF-IDF and BM25: text-based, non-trainable retrieval methods \citep{robertson2009probabilistic} which are strong baselines in common retrieval benchmarks \citep{thakur2021beir}.  This suggests that \method is able to learn more than just superficial text similarity.
In all these experiments we used the same evaluation protocol (following Algorithm~\ref{alg:evaluate_psm_on_proof_rate_benchmark}) and computational budget of $1000$ as detailed in Appendix \ref{app:computational_budget}.


Interestingly, retrieval based on the generic OpenAI embeddings \citep{neelakantan2022text} (specifically: text-embedding-ada-002) yields reasonable performance comparable to Sledgehammer. This confirms the potential of neural premise selection to replace traditional symbolic methods. There is, however, a large gap to match Magnushammer. This shows that contrastive fine-tuning on our dataset provides non-trivial gains and supports our hypothesis that Magnushammer learns more than just mere textual similarity exploited by the general purpose method.

\paragraph{Performance on the multi-step task}\label{sec:main_result_multi}
Neural theorem provers utilize language models to generate proof steps, following the approach proposed by \citet{polu2020generative}. This allows for the creation of more complex, multi-step proofs. The proof generation involves sampling a proof step from the language model, verifying it, and repeating this process until the proof is closed or the computational budget is exceeded. The best-first search algorithm is often used to explore the most promising proof steps.

Thor \citep{jiang2022thor} augments neural theorem provers with premise-selection capabilities. To this end, Thor allows the model to generate proof steps using Sledgehammer, which we replace with \method{} (see Appendix \ref{app:thor} for details). Thor + \method{} establishes a new state of the art on the PISA benchmark ($71.0\%$ vs. $57.0\%$).
On miniF2F, our method also significantly outperforms Thor and achieves results competitive with the current state of the art. In these experiments, we give \method a computational budget of $200$.

It is important to note that other theorem-proving approaches in the multi-step
section of Table \ref{tab:minif2f_results} require much larger language models:
for Thor it is $700$M non-embedding parameters; DSP (Draft, Sketch, and Prove) by \citet{jiang2022draft} uses
Minerva model \citep{lewkowycz2022solving} with $62$B parameters. Moreover, these other
approaches rely on ideas orthogonal to premise selection. Specifically, Thor +
auto \citep{autoform_wu} proposes a variation of Thor, involving expert
iteration on auto-formalized data. DSP involves creating a high-level outline
of a proof and uses Sledgehammer to solve the low-level subproblems. We
hypothesize that both methods would perform even better when combined with
\method{}.

\subsubsection{Scaling computational budget} \label{sec:budget_experiments}

In this section, we discuss how the quality of premise selection methods varies with the computational budget available during evaluation. Figure \ref{fig:compute_cost_ablation} shows the results, and
the definition of the compute budget is provided in Section \ref{sec:compute_budget_definition}.
Notably, \method{} outperforms Sledgehammer even with very limited computational resources, and it scales well, particularly within the medium budget range.

For \method and BM25, we use Algorithm \ref{alg:evaluate_psm_on_proof_rate_benchmark}
 (Appendix \ref{app:eval_on_theorems})
in various configurations (i.e., settings of $\mathcal{T}$ and $K$). We  start with one tactic, $\mathcal{T} = \{\mathtt{smt}\}$, and $K=[2^7]$, which yields $C = 2$ (recall that $T=2$ s). We then gradually add more tactics to $\mathcal{T}$ and more values to $K$. The final setup uses $|\mathcal{T}| = 36$ and $K$ containing all powers of $2$, from $2^0$ up to $2^{10}$, which yields $C\approx800$. Details are provided in Appendix \ref{app:eval_on_theorems}. For Sledgehammer, we scale the timeout parameter $T$ up to $80$ s.

\subsection{Impact of training data}\label{sec:dataset_size_ablation}

We study how the amount and type of data impact the proof success rate by comparing \textsc{HPL} and \textsc{MAPL} datasets. For this comparison, we used models with $38$M non-embedding parameters and a computational budget of $800$.

\begin{figure}
\centering
\begin{subfigure}{0.49\linewidth}
\includegraphics[width=\linewidth,clip]{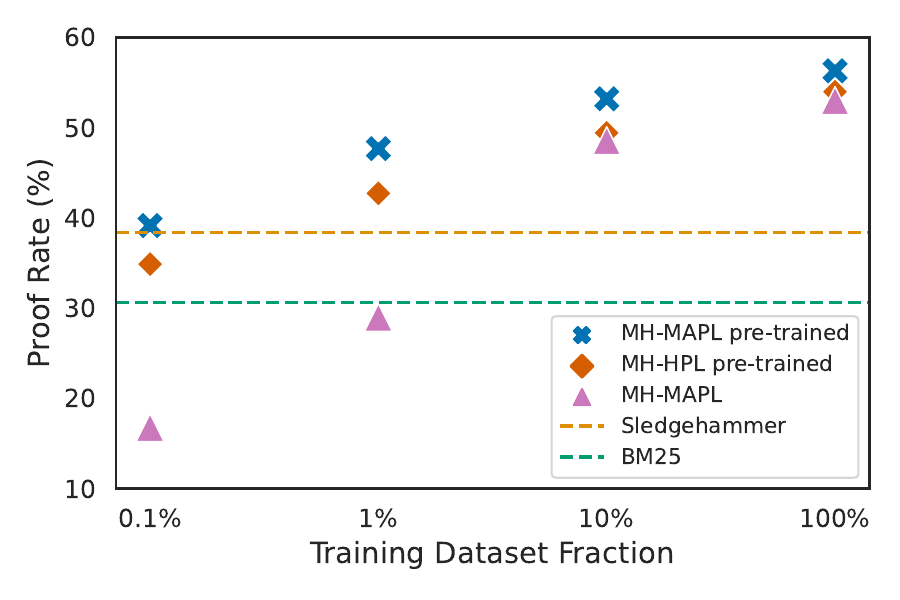}
    \caption{We randomly sample fractions of \textsc{MAPL} or \textsc{HPL} datasets and use them for training \method. Even $0.1\%$ of the \textsc{MAPL} dataset allows pre-trained \method to outperform the Sledgehammer and BM25 baselines. See Table \ref{tab:traning_data_impact} for numerical data.}
    \label{fig:dataset_size_ablation}
\end{subfigure}
\hfill
\begin{subfigure}{0.49\linewidth}
\includegraphics[width=\linewidth,trim={0 5pt 0 0},clip]{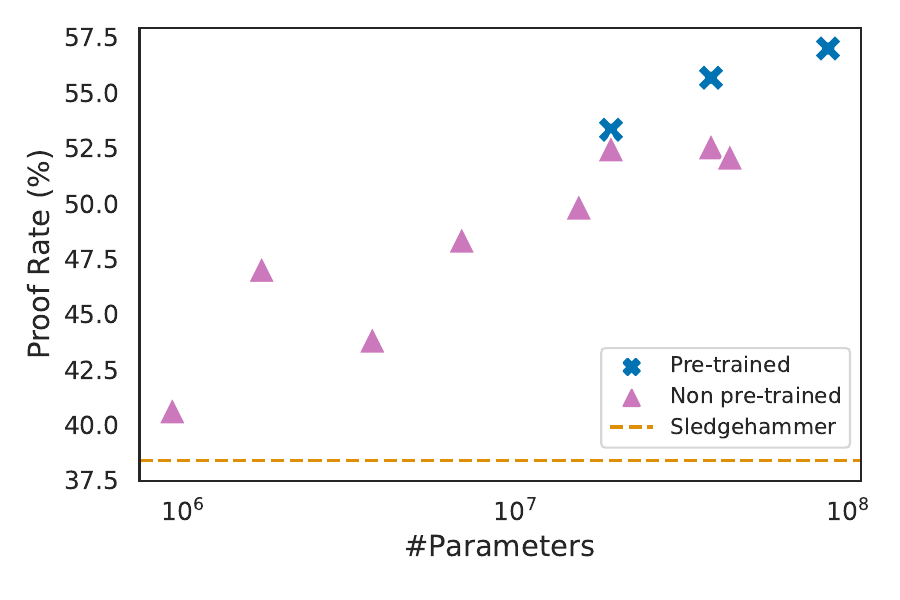}
    \caption{We train \method of different sizes. Even with a one-layer transformer, \method outperforms Sledgehammer.
    We observe consistent performance gains with increasing model sizes. Pre-trained models perform better. See Table \ref{tab:model_size_ablation} for numerical data.}
    \label{fig:model_size_ablation}
\end{subfigure}
\caption{
Impacts of the training data quantity and the model parameters on the proof rate. The vertical axis is the proof rate in percentage. In Subfigure~\ref{fig:dataset_size_ablation}, the horizontal axis is the fraction of training dataset used and in Subfigure~\ref{fig:model_size_ablation} it is the number of parameters in the model.
}
\end{figure}

\paragraph{Dataset size} Our method is data-efficient: see Figure~\ref{fig:dataset_size_ablation}. We observe that \method fine-tuned on only $0.1\%$ of MAPL -- equivalent to approximately $4$K samples -- is already able to outperform Sledgehammer. This indicates that when starting from a pre-trained model, \method is a promising approach for addressing premise selection in theorem-proving environments with limited training data. The effect of pre-training diminishes as the amount of training data increases.

\paragraph{Dataset type}
Fine-tuning on \textsc{MAPL} or \textsc{HPL} leads to subtle differences ($56.3\%$ vs. $54.0\%$ when the whole datasets are used). This outcome may be attributed to the impact of model pre-training and the fact that the \textsc{HPL} dataset is rich enough to obtain good performance on the PISA benchmark (as observed in the previous paragraph). We speculate that the bigger \textsc{MAPL} dataset might be essential for future harder benchmarks and scaling up the model size.

\subsection{Ablations}\label{sec:model_size_ablation}

We use models trained on the \textsc{MAPL} dataset and evaluate them with a computational budget of $800$.
To study how the performance of our method depends on the model size, we vary the number of layers $L$ and embedding dimension $D$.
A positive correlation between the model size and the proof rate is shown in
Figure \ref{fig:model_size_ablation}. We observe that even a tiny model with
$920$K parameters ($L=1, D=256$) outperforms Sledgehammer ($40.7\%$ vs.
$38.3\%$). We also note the benefit of pre-training and that scaling the number
of layers is more beneficial than scaling the embedding dimension. Details
can be found in Appendix
\ref{app:model_architecture}. The impact of re-ranking is studied in Appendix
\ref{sec:reranking_ablation}.




\section{Related work}\label{sec:related_work}

Premise selection becomes a crucial task whenever proving theorems
automatically within a large formal library. Moreover, this task has several
unique aspects that are challenging from the perspective of learning-based
approaches. Therefore, there exist multiple works that tackle learning premise
selection (either explicitly or implicitly) applying various methods focusing
on different aspects.

Many works employ classical machine learning like Bayesian and kernel
methods~\citep{kuhlwein2012overview, alama},
$k$-NN~\citep{fact-selector-isabelle}, or decision trees
\citep{atpboost,pamper,lean-premsel}. The common weakness of these approaches is
the necessity of using hand-engineered features, whereas faster, simpler training is an advantage.

\citet{alemi2016deepmath} were the first to apply deep learning to premise selection,
thus dispensing with the hand-designed features completely. Their approach was evaluated
in an automated theorem proving setting and not in a proof assistant, as is Magnushammer.
They also implicitly learn embeddings of conjectures and premises, which are concatenated and
passed through a shallow network, whereas the training signal comes from the logistic loss.
In contrast, Magnushammer demonstrated the strength of training with the
contrastive loss, where the obtained embeddings just need to be passed through a
simple cosine similarity measure to provide high-quality rankings.

Most of the methods explicitly targeting the premise selection problem
(including this work) retrieve a \textit{ranking} of independently treated premises.
In contrast, \citet{piotrowski2020stateful} aimed at modelling the implicit
dependencies \textit{between} the premises and used LSTM-based language models
to produce structured sequences of premises. However, the premises were treated
there as opaque tokens, not giving the neural model the ability to inspect the
statements of the premises.

Effective deep learning approaches often leverage the explicit structure of
mathematical expressions using graph neural networks~\citep{wang2017premise,
paliwal2019graph, goertzel2022isabelle}.
Our work uses the transformer architecture \citep{vaswani2017attention}, which
is highly scalable and capable of producing powerful representations of raw text
data.

Pre-trained transformer language models have been applied to various aspects of
theorem proving, including autoformalization~\citep{autoform_wu,
jiang2022draft}, conjecturing~\citep{conj}, and tactic prediction / proof step
search~\citep{yang2019learning, polu2020generative, han2021proof,
lample2022hyper, curr}. The works from the last category often implicitly deal with
premise selection by treating premises as names / tokens to be generated and not
inspecting their statements. The application of generative language models to
statement-aware premise selection has been limited, as the length of the
possible premises often greatly exceeds the context of several thousand tokens
that the models are designed to handle. Thor~\citep{jiang2022thor} circumvents
the difficulty of premise selection by invoking Sledgehammer.  In contrast,
\method \textit{retrieves} rather than \textit{generates} to overcome the context length
limitation.  Therefore it can be used in tandem with other models~(its
combination with Thor is demonstrated in Section~\ref{sec:experiments}).

Batch-contrastive learning is widely used in speech
\citep{oord2018representation}, text \citep{contriver}, image \citep{simclr}
and image-text \citep{radford2021learning} representation learning. These
methods have proven effective despite the possibility of false negatives
occurring in contrastive batches~\citep{robinson2021contrastive}. The \methods
phase of our premise selection model relies on in-batch negative examples to
train the retriever, similar to HOList~\citep{bansal2019holist} and
Contriever~\citep{contriver}. Like HOList, we mine additional negatives, which
we found crucial for performance. The \methode stage closely resembles
\citep{nogueira2019passage}, but instead of using BM25, we jointly train
retrieval and re-ranking, utilizing premises retrieved by \methods as hard
negatives for \methode training.
\citet{han_informal_premise_selection} use contrastive learning in informal
premise selection.
Concurrently to our work, \citet{yang2023leandojo} develop a premise selection
method for Lean also using contrastive learning in a way similar to our \methods
method, but without the \methode stage.

There are multiple lines of work considering datasets based on formal theorem proving.
These include benchmarks like ProofNet~\citep{azerbayev2022proofnet} for Lean, and miniF2F~\citep{zheng2021minif2f} that supports multiple ITPs.
These datasets only focus on evaluation, not providing data for training the models. Another line of research focuses on benchmarking machine learning models' reasoning capabilities while also providing training data \citep{bansal2019holist, li2020isarstep, han2021proof}. Existing public datasets for premise selection include the ones introduced in \citep{alama,piotrowski2020stateful}. In comparison to these works, we publish the data in high-level, textual format, as seen in Isabelle, instead of low-level, structured languages such as TPTP \citep{tptp}. 

There exists a rich body of work developing complex hammers systems for
different proof assistants
\citep{Paulson2012ThreeYO, holyhammer, hol4hammer, coqhammer}. Unlike the
traditional hammers, our method does not depend on external ATPs and requires
little domain-specific knowledge.

\section{Limitations and future work}
\paragraph{Other proof assistants} \method treats proof states and premises as text and makes no assumptions about their structure. As such, no feature engineering is needed to apply it to other proof assistants. We conjecture that \method can prove effective in other environments because it is agnostic to the logic or type system used.
We plan to evaluate Magnushammer in Lean proof assistant on ProofNet~\citep{azerbayev2022proofnet} and miniF2F~\citep{zheng2021minif2f}  benchmarks, using the recently published LeanDojo toolkit~\citep{yang2023leandojo} that also provides baselines for comparison.

\paragraph{Richer proof and premise representations}  \method utilizes the textual representation of the proof state given by Isabelle. This representation, however, does not provide complete semantic information about the referenced objects. Including function definitions and object types in the proof state representation might further improve performance.

\paragraph{Modelling full proof steps} Combining language models with external premise selection tools significantly improves their theorem-proving performance, as demonstrated by \citet{jiang2022thor} and our work. A natural step would be to further integrate premise selection with language models into a single model capable of generating proof steps containing relevant retrieved premises. A proof of concept of this idea was explored by \citet{fps_with_lms}. This would also allow to model existing implicit dependencies between returned premises, which was shown beneficial by \citet{piotrowski2020stateful}.
We believe that recent advances in retrieval-augmented language models \citep{wu2022memorizing, borgeaud2022improving} could facilitate progress in this direction.


\section*{Reproducibility statement}

The data that were used for pre-training of the backbone transformer model of \method are freely available under this link:
\url{https://pile.eleuther.ai/}

The Isabelle data used in training for the down-stream tasks are available
under this link: \\
\url{https://huggingface.co/datasets/Simontwice/premise_selection_in_isabelle}

The benchmarks used for evaluation of Magnushammer are freely available on GitHub:
\begin{compactitem}
\item miniF2F: \url{https://github.com/openai/miniF2F}
\item PISA: \url{https://github.com/albertqjiang/Portal-to-ISAbelle}
\end{compactitem}

PISA also implements the interface for interacting with Isabelle that we used in our experiments.

Appendix \ref{app:sledgehammer} specifies the setup of Sledgehammer that we used in our comparisons. Appendices \ref{sec:selectAndExpandAppendix} and \ref{app:trainingDetails} detail the shape of the transformer architecture used, define the loss functions applied in the \methods and \methode stages, specify the hyperparameters used in pre-training and training for our down-stream tasks, and disclose the hardware used for training.
Appendix \ref{app:eval_on_theorems} details the setup for evaluation of \method in Isabelle, in particular the list of tactics applied on top of the \method's premise selection.

\newpage
\bibliography{bibliography}
\bibliographystyle{plainnat}
\newpage
\appendix
\onecolumn
\appendix
\section*{Appendix}

\setcounter{figure}{0}
\renewcommand{\thefigure}{A.\arabic{figure}}

\section{Isabelle environment}\label{sec:isabelle}
This section contains visual examples of proofs in Isabelle and provides some configuration details of the environment.

\subsection{Visualization of the Isabelle environment}
\label{app:visualization_of_isa_env}
\begin{figure}[H]
    \centering
    \includegraphics[width=0.85\textwidth]{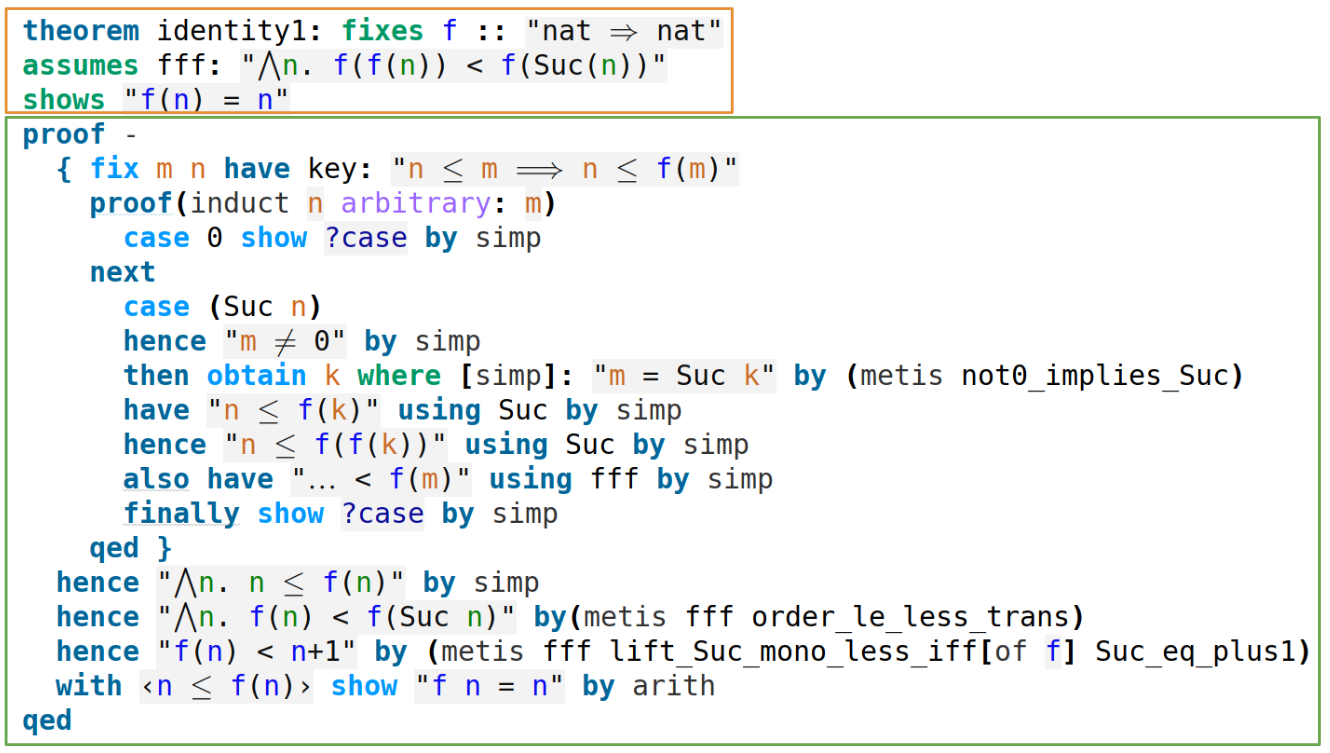}
    \caption{An example theorem in Isabelle. The statement is highlighted in the orange frame and the body of the proof is in the green frame. In this proof, most of the lines contain two consecutive steps: the first formulates a new proposition, and the second proves it. See a detailed analysis of the line 8 of the proof in Figure \ref{issa2} below.
    }
    \label{issa1}
\end{figure}

Figure \ref{issa1} shows an example theorem and its proof, as seen in Isabelle's most popular IDE, jEdit. The theorem comes from an entry to the Archive of Formal Proofs -- \textit{Fun With Functions} \citep{FunWithFunctions-AFP}. It states that any mapping $f$ from the set of natural numbers to itself that satisfies $f(f(n))<f(n+1)$ must be the identity function. The proof starts with a simple induction and then refines the result to arrive at the thesis. This problem was included in Terence Tao's booklet \textit{Solving Mathematical Problems} \citep{tao_2010}.

\begin{figure}[H]
    \centering
    \includegraphics[width=0.75\textwidth]{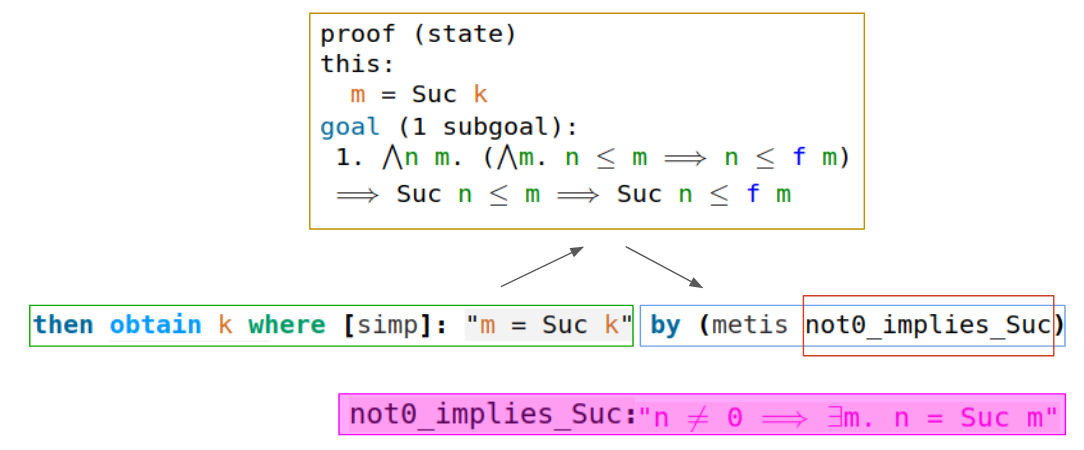}
    \caption{The line is broken down into two steps: the first one (green frame) includes the proposition (since $m$ is natural and positive, it must have a predecessor $k$) and the second (blue frame) proves it using the tactic \texttt{metis} with premise \texttt{not0\_implies\_Suc}, that states that a nonnegative natural number is a successor of some other natural number. The used premise is a fact which is already defined in the lemma library. The proof state resulting from the first step is in the yellow frame. The full premise statement is highlighted in pink.}
    \label{issa2}
\end{figure}

\subsection{Alternative proof step generation with Sledgehammer}
\label{app:alternative_proof_step_generation_with_sh}
\begin{figure}[ht]
    \centering
    \includegraphics[width=0.9\textwidth]{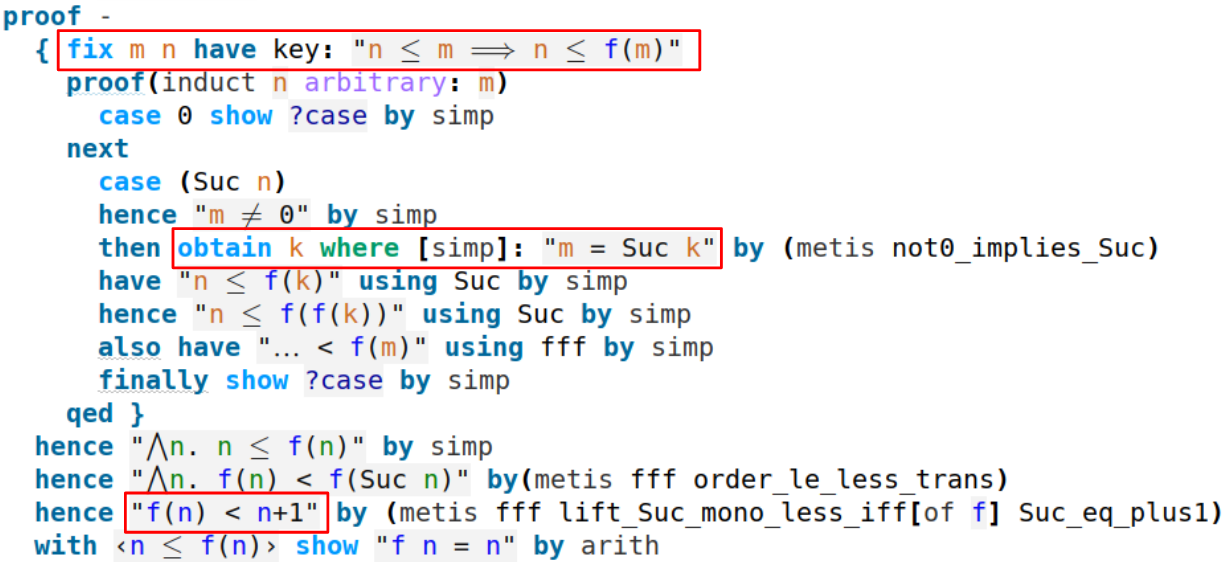}
    \caption{Example intermediate propositions highlighted in red. Note:  not all propositions were highlighted.
    }\label{issa3}
\end{figure}

This section describes how to generate alternative proof steps using Sledgehammer which we do to obtain datasets described in Section \ref{sec:datasets}. First, we find all intermediate propositions within the proof (they can be nested) and try to replace the proof of the proposition with a Sledgehammer step. If successful, we record such a step in the dataset and proceed with both the original and the alternative proof. Figure \ref{issa3} provides a visual example of the aforementioned propositions.

\subsection{Example of a proof with tactics requiring premises}\label{app:proof_with_premises}
\begin{figure}[H]
    \centering
    \includegraphics[width=0.8\textwidth]{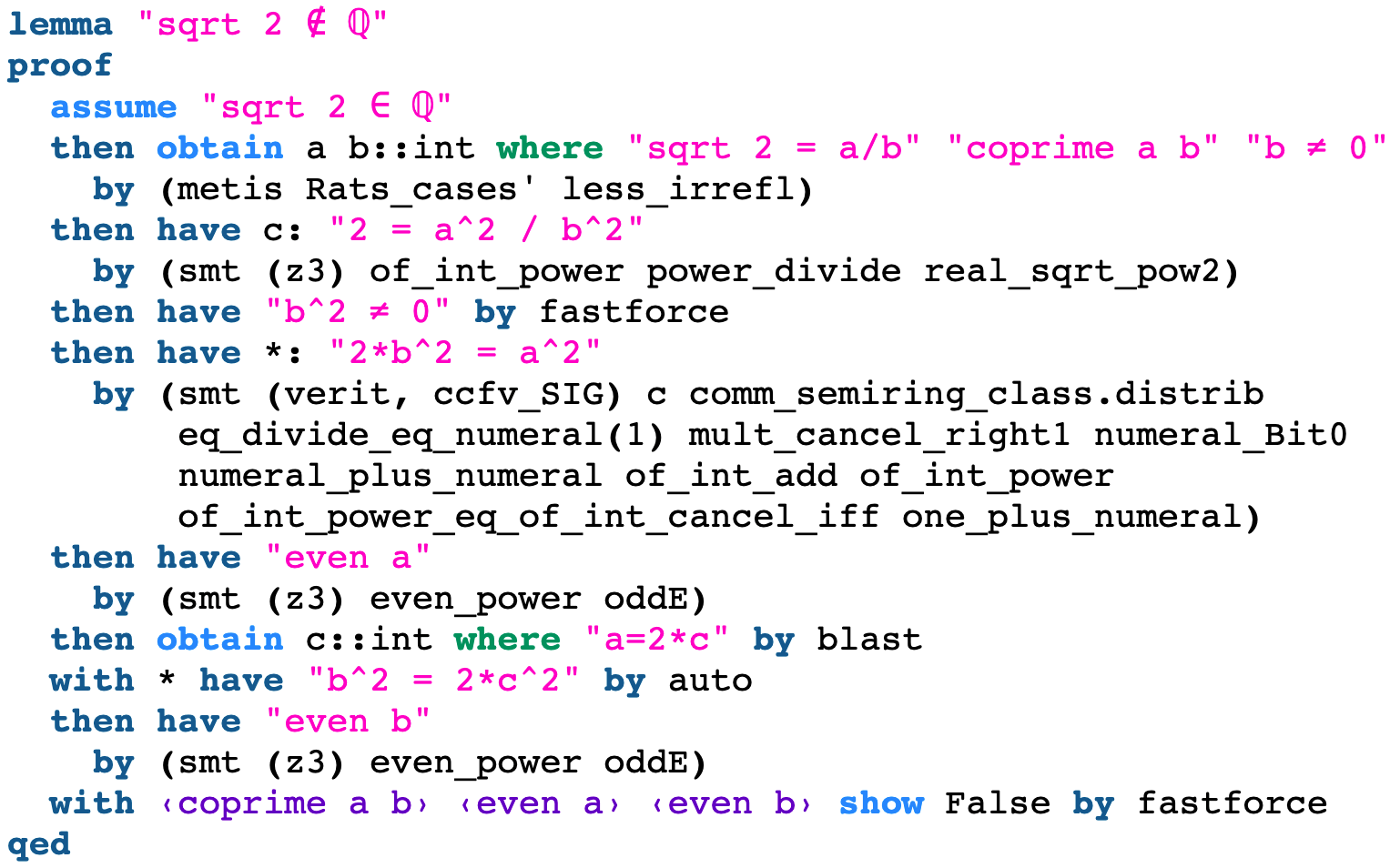}
    \caption{A proof of $\sqrt{2} \notin \mathbb{Q}$ \citep[Figure~1]{jiang2022thor}. The steps containing \texttt{metis, smt, fastforce, blast, auto, fastforce} are examples of steps using premises. For instance, one such proof step is \texttt{by (metis Rats\_cases' less\_irrefl)}. This step invokes \texttt{metis} and provides two premises as arguments, namely \texttt{Rats\_cases'} and \texttt{less\_irrefl}.
    }
    \label{fig:isa_proof_example}
\end{figure}

Figure \ref{fig:isa_proof_example} contains a multi-step proof of the irrationality of $\sqrt{2}$ written in Isabelle. The proof contains multiple usages of tactics that require premises.

\subsection{Sledgehammer setup} \label{app:sledgehammer}
We set up Sledgehammer in Isabelle 2021-1, following the configuration used by \cite{jiang2022thor}. We run Sledgehammer using different sets of settings and calculate the total proof rate by taking the union of problems solved by each run. The Sledgehammer timeout is set to default $30$ seconds. We use only on-machine automated theorem provers (same as Isabelle environment), so external provers used by Sledgehammer are the following: Z3, SPASS, Vampire, CVC4, and E.

In our calculation of the Sledgehammer computation budget, see Section \ref{sec:exp_details}, we assume $S=10$ 'CPU cores.' We run our experiments on machines with $96$ CPU cores, making the assumption realistic. Moreover, we emphasize that the performance gap between \method{} and Sledgehammer is large enough that altering the value of $S$, e.g., to an unrealistic level $S=1$, would not qualitatively change conclusions.

\section{Details of \method}\label{sec:selectAndExpandAppendix}

\subsection{\methods stage}
\methods{} stage is trained using the InfoNCE loss \cite{oord2018representation} defined as:
\[
\mathcal{L}\left(q, k_{+}\right)=-\frac{\exp \left(s\left(q, k_{+}\right) / \tau\right)}{\exp \left(s\left(q, k_{+}\right) / \tau\right)+\sum_{i=1}^K \exp \left(s\left(q, k_i\right) / \tau\right)},
\]
where $q$ is a query (a proof state), $k_{+}$ is a positive premise (a ground truth from the dataset), $k_i$ are negative premises. We define $s$ as cosine similarity between proof state and premise embeddings; $\tau>0$ is a non-trainable temperature parameter. We list our hyperparameter choices in section \ref{sec:hyperparameter}.

\subsection{\methode stage}\label{app:expand}
Premise retrieval task can be cast as binary classification, trying to determine if a given pair $(\mathtt{proof\_state}, \mathtt{premise})$ is relevant. Applying classification to each pair is computationally infeasible, however, it could be used to \textit{re-rank} a small set of premises retrieved by \methods.
Namely, we use the following cross-entropy loss:
\[
\mathcal{L} = - \sum_{p \in \mathcal{P}} \log \mathtt{score}(p) - \sum_{p \notin \mathcal{N}} \log (1 - \mathtt{score}(p)),
\]
where $\mathtt{score}(p)$ is the output of the \methode{} part of the model (see "Sigmoid" in Figure \ref{fig:select_expand_overview}) for a given $p = (\mathtt{proof\_state}, \mathtt{premise})$ pair. Typically, we sample a batch of $16$ positive pairs  $\mathcal{P}$ from the dataset. For each such pair $(\mathtt{proof\_state}, \mathtt{premise})$ $15$ negatives are constructed from the most likely false positives returned by \methods{}. Specifically, negative premises $\mathcal{\mathcal{M}}$, which are facts that were never used as a premise for $\mathtt{proof\_state}$, are first chosen. Then, the top $1024$ of $\mathcal{M}$ according to \methods{} are selected, and $15$ are sampled from them to construct negative pairs, which are included in $\mathcal{N}$.




\subsection{\method}
\label{appendix:method}
We train \method as two separate tasks alternating update steps as presented in Algorithm \ref{alg:train_select_expand}. Note that the backbone of the architecture is shared between \methods{} and \methode{}, thus such multi-task training is potentially more effective than having two separate models. Calculation of the negative premises for \methods{} is costly, thus for efficiency reasons we recalculate the top $1024$ premises, see Section \ref{app:expand}, every $T = 1000$ steps in the $\mathtt{recompute\_negatives\_for\_rerank}$ function, as outlined in the Algorithm \ref{alg:train_select_expand}.

\begin{algorithm}[h]
\caption{\method training.}\label{alg:train_select_expand}
\begin{algorithmic}[1]
\Require
\Statex $\theta$ \Comment{\makebox[320pt][l]{initial trainable parameters}}
\Statex $D$ \Comment{\makebox[320pt][l]{premise dataset}}
\Statex $T$ \Comment{\makebox[320pt][l]{interval for updating rerank dataset}}
\State $D\mathtt{_{rerank}} \gets$  {$\mathtt{recompute\_negatives\_for\_rerank}$}{$(\theta, D)$}
\State $\mathtt{step} = 0$
\While{$\mathtt{step} < \mathtt{num\_train\_steps}$}
\State $\mathtt{batch\_select} \gets D\mathtt{.sample()}$
\State $\mathtt{\theta} \gets$ {train\_step}{$\mathtt{(\theta, batch\_select)}$}
\State $\mathtt{batch\_rerank} \gets D\mathtt{_{rerank}.sample()}$
\State $\mathtt{\theta} \gets$  {train\_step}{$\mathtt{(\theta, batch\_rerank)}$}
\State $\mathtt{step} \gets \mathtt{step} + 1$
\If{$\mathtt{step}\mod T = 0$}
\State $D\mathtt{_{rerank}} \gets$  {$\mathtt{recompute\_negatives\_for\_rerank}$}{$(\theta, D)$}
\EndIf
\EndWhile
\end{algorithmic}
\end{algorithm}



\section{Training details} \label{app:trainingDetails}

\subsection{Model architecture}\label{app:model_architecture}

We use a decoder-only transformer architecture, following the setup from \cite{gpt-j} and using rotary position embedding by \cite{su2021roformer}, a variation of relative positional encoding. The feedforward dimension in the transformer block is set to $4\times D$ where $D$ denotes embedding dimension, and the number of attention heads is $H=D/64$. Our $38$M model has $L=12$ layers and an embedding dimension of $D=512$. The larger $86$M model consists of $L=12$ layers and has $D=768$. For all the models, we use the original GPT-2 tokenizer \citep{radford2019language}.

In \methods, we append a specialized token at the end of the sequence to compute the embedding for a proof state and linearly project its embedding. Premises are embedded analogously. Similarly to \cite{radford2021learning} that train separate projections for images and captions, we train separate proof state and premise projections and share the transformer backbone (see Figure \ref{fig:select_expand_overview}). Analogously for \methode{}, we compute the relevance score by taking the embedding of the last token and then projecting it to a scalar value.

\newcommand{\expnumber}[2]{{#1}\mathrm{e}{#2}}
\subsection{Hyperparameter setup}\label{sec:hyperparameter}

We performed the following hyperparameter sweeps. We note that we have not observed significant differences between obtained results.
\begin{itemize}
    \item Learning rate: $\{\expnumber{1}{-4}, \expnumber{2}{-4}, \expnumber{3}{-4}, \expnumber{5}{-4} \}$, chosen: $\expnumber{2}{-4}$
    \item Dropout: $\{0.0, 0.05, 0.1, 0.2\}$, chosen: $0.1$
    \item Weight decay: $\{0.02, 0.05, 0.1\}$, chosen: $0.02$
    \item Batch size $N$ in \methods: $\{ 128, 256, 512\}$, chosen: $256$
    \item Number of negatives $M$ in \methods: $\{ 0, 256, 768, 1536\}$, chosen: $768$
    \item Temperature for InfoNCE loss in \methods: $\{0.05, 0.07, 0.2, 1\}$, chosen: $0.07$
    \item Batch size for \methode: $\{ 16, 32, 64\}$, chosen $64$
    \item Number of negatives per proof state $\mathcal{M}$ in \methode: $\{7, 15\}$, chosen: $15$.
\end{itemize}


\subsection{Pre-training on language modeling}\label{app:pre-training}
Pre-training has been shown to dramatically increase the capabilities and performance of decoder-only models on tasks other than language modeling \citep{howard2018universal}. Motivated by that, we pre-train our models on GitHub and arXiv subsets of the Pile \citep{gao2021pile}. The models are trained for $1$M steps, with a context length of $2048$. Global batch size is set to $32$ sequences giving a total number of $65536$ tokens per batch. Dropout is disabled, and weight decay is set to $0.02$. The learning rate increases linearly from $0$ to $0.0003$ for the first $10 000$ steps, and then the cosine schedule is applied to decrease its value gradually.

\subsection{Fine-tuning for downstream tasks}\label{app:heads}\label{app:fine_tuning_on_proof_generation}
We train \method by taking a pre-trained language model, removing its language modeling head, and attaching three linear projections heads -- one projection for proof state embedding, another one for premise embedding, and the last one for producing relevance score for \methode, as depicted in Figure \ref{fig:select_expand_overview} and described in Section \ref{app:model_architecture}. For the proof step generation task, we fine-tune our language models by applying the algorithm used to train Thor \citep{jiang2022thor}.



\subsection{Impact of re-ranking}\label{sec:reranking_ablation}
We find that the \methods-only method, i.e., \method{} without the \methode{} phase, already significantly outperforms Sledgehammer. Tested on the $38$M model, it achieves a $54.2\%$ proof rate comparable to $56.3\%$ obtained by \method{}.
\methods-only mode is a computationally appealing alternative, as it only needs a single forward pass to embed the current proof state (the setting used recently by \citet{yang2023leandojo}.) Premise embeddings can be pre-computed and cached, allowing inference on the CPU without the need for GPU or TPU accelerators.

\subsection{Hardware}
We gratefully acknowledge that our research was supported with Cloud TPUs from Google's TPU Research Cloud (TRC).
We use TPU virtual machines from the Google Cloud Platform (GCP) for all stages: pre-training, fine-tuning, and evaluation. Each TPU virtual machine has 8 TPU v3 cores, 96 CPU cores, and over 300GB of RAM. TPU v3 cores have around 16GB of memory each. The Isabelle environment is set to have access to 32 CPU cores.

\section{\method evaluation} \label{app:eval_on_theorems}

In Algorithm \ref{alg:evaluate_psm_on_proof_rate_benchmark} we outline our evaluation method described in Section \ref{sec:exp_details}. To generate proof steps, we use the following tactics:
\texttt{smt}, \texttt{metis}, \texttt{auto}, \texttt{simp}, \texttt{blast}, \texttt{meson}, \texttt{force}, \texttt{eval}, \texttt{presburger}, \texttt{linarith}. Algorithm \ref{alg:evaluate_psm_on_proof_rate_benchmark} is also used to evaluate BM25, where we select $\mathtt{top\_premises}$ with this retrieval method instead of \method{}.

\subsection{Computational budget}\label{app:computational_budget}
For our main result (Section \ref{sec:main_result_single}), we allocate the computational budget of $1000$ as follows: apart from the powers of two from $2^0$ to $2^{10}$, we also try the following $k$ values: $\left[48, 96, 192\right]$, which in total gives $14$ values. With each of these $k$ values, $36$ tactics are used with timeout $T=2$, yielding $C \approx 1000$.

For the ablation studies, we only use powers of two from $2^0$ to $2^{10}$, and the same set of $36$ tactics, which gives $C \approx 800$.

\begin{algorithm}
   \caption{\method evaluation in ITP environment.}
   \label{alg:evaluate_psm_on_proof_rate_benchmark}
\begin{algorithmic}[1]
\Require
\Statex $\mathtt{theorem}$ \Comment{theorem to prove}
\Statex $\mathtt{premsel\_model}$ \Comment{\method's premise selection model}
\Statex $K_{S}$ \Comment{number of premises to retrieve with \methods{}}
\Statex $K_{R}$ \Comment{number of premises to retrieve with \methode{}}
\Statex $\mathtt{premises}$ \Comment{available premises}
\Statex $\mathtt{top\_k\_premises\_to\_try}$ \Comment{list with the number of top premises to generate steps with}
\Statex $\mathtt{tactics\_to\_try}$ \Comment{list of tactics to generate steps with}
\Statex $\mathtt{env}$ \Comment{ITP environment (e.g., Isabelle)}
\State $\mathtt{proof\_state} \gets \mathtt{init\_problem}(\mathtt{env}, \mathtt{theorem})$
\Comment{initialize problem}
\State $\mathtt{top\_premises} \gets$ $\mathtt{premsel\_model}({\mathtt{proof\_state}}, \mathtt{premises}, K_{S}, K_{R})$
\Comment{get top premises}
\State $\mathtt{steps}$ = []
\Comment{generate proof steps combining of tactics and top $k$ premises}
\For{$\mathtt{k}$ \textbf{ in } $\mathtt{top\_k\_premises\_to\_try}$}
\State $\mathtt{top\_k\_premises} \gets$ $\mathtt{top\_premises[:k]}$
\State $\mathtt{new\_steps} \gets \mathtt{generate\_steps}(\mathtt{tactics\_to\_try}, \mathtt{top\_k\_premises})$
\State $\mathtt{steps.extend}(\mathtt{new\_steps})$
\EndFor
\State $\mathtt{solved} \gets \mathtt{try\_steps}({\mathtt{env, steps}})$
\Comment{evaluate generated proof steps in the ITP's environment}
\State \textbf{return} $\mathtt{solved}$

\end{algorithmic}
\end{algorithm}

\subsection{Thor + \method}\label{app:thor}
To generate more complex proofs we combine Thor \citep{jiang2022thor} with \method as introduced in multi-step setting in Section \ref{sec:main_result_multi}.


Firstly, we follow the procedure described in \cite{jiang2022thor} to pre-process training data and fine-tune our pre-trained language model for the proof generation task (pre-training details can be found in Appendix \ref{app:pre-training}). During the evaluation, when the language model generates the \texttt{<hammer>} token, we call our method instead of Sledgehammer. More specifically, we use an augmented Algorithm \ref{alg:evaluate_psm_on_proof_rate_benchmark} that returns the proof states resulting from applying the steps (instead of returning binary information on whether any of the steps closed the proof). We then pick at most s = 2 states among these and add them to the BFS queue.



We assign the same computational budget as proposed in Thor, with the only difference being
that each \texttt{proof\_step} has a timeout limit of $2$ s (instead of $10$ s), which we found to perform better in our setup.
The search is terminated if and only if one of the following scenarios happens: (1) a valid proof has been found for the theorem; (2) the language model is queried 300 times; (3) a wall-time timeout of $500$ s has been reached (assuming parallel execution of \method steps); (4) the queue is empty but the theorem is not proved.
We keep the same maximum length of the queue equal to $32$.

\section{Additional experimental results} \label{app:additionalResults}

\subsection{Supplemental details}\label{app:supplementary_details}

We provide additional details for our main experiments and ablations.

\begin{table}[h]
\begin{center}
\caption{Relation between the training data and the proof rate discussed in Section \ref{sec:dataset_size_ablation} and Figure \ref{fig:dataset_size_ablation}.}
\label{tab:traning_data_impact}
\begin{tabular}{ccccc}
\toprule
Dataset & Fraction & Pre-trained & Proof rate~(\%) \\
\midrule
MAPL & $0.1\%$ & Yes & $39.2$ \\
HPL & $0.1\%$ & Yes & $34.9$ \\
MAPL & $0.1\%$ & No & $16.8$ \\
\midrule
MAPL & $1\%$ & Yes & $47.7$ \\
HPL & $1\%$ & Yes & $42.7$ \\
MAPL & $1\%$ & No & $29.0$ \\
\midrule
MAPL & $10\%$ & Yes & $53.2$ \\
HPL & $10\%$ & Yes & $49.4$ \\
MAPL & $10\%$ & No & $48.5$ \\
\midrule
MAPL & $100\%$ & Yes & $56.3$ \\
HPL & $100\%$ & Yes & $54.0$ \\
MAPL & $100\%$ & No & $53.0$ \\
\bottomrule
\end{tabular}
\label{tab:dataset_impact}
\end{center}
\end{table}


\begin{table}[h]
\begin{center}
\caption{Proof rate on PISA for different models discussed in Section \ref{sec:model_size_ablation} and Figure \ref{fig:model_size_ablation}. We vary the number of layers $L$ and the embedding dimension $D$ of the Transformer model.}\label{tab:model_size_ablation}
\begin{tabular}{ccccc}
\toprule
Transformer $(L,D)$ & \#Parameters & Pre-trained  & Proof rate~(\%) \\
\midrule
$(1,256)$ & $920$K & No & $40.7$ \\
$(1,512)$ & $3.7$M & No & $43.9$ \\
$(2,256)$ & $1.7$M & No & $47.0$ \\
$(2,512)$ & $6.8$M & No & $48.4$ \\
$(2,768)$ & $15.4$M & No & $49.9$ \\
$(6,512)$ & $19.2$M & No & $52.5$ \\
$(6,512)$ & $19.2$M & Yes & $53.3$ \\
$(6,768)$  & $43.7$M & No & $52.1$ \\
$(12,512)$ & $38.3$M & No & $52.6$ \\
$(12,512)$ & $38.3$M & Yes & $56.3$ \\
$(12,768)$ & $86.2$M & Yes & $57.0$ \\
\bottomrule
\end{tabular}
\label{tab:app_model_size}
\end{center}
\end{table}

\subsection{Step tactic prompt}
\label{app:step_tactic_prompt}

\begin{figure}[ht]
    \begin{center}
    \centering
\centerline{\includegraphics[scale=0.6]{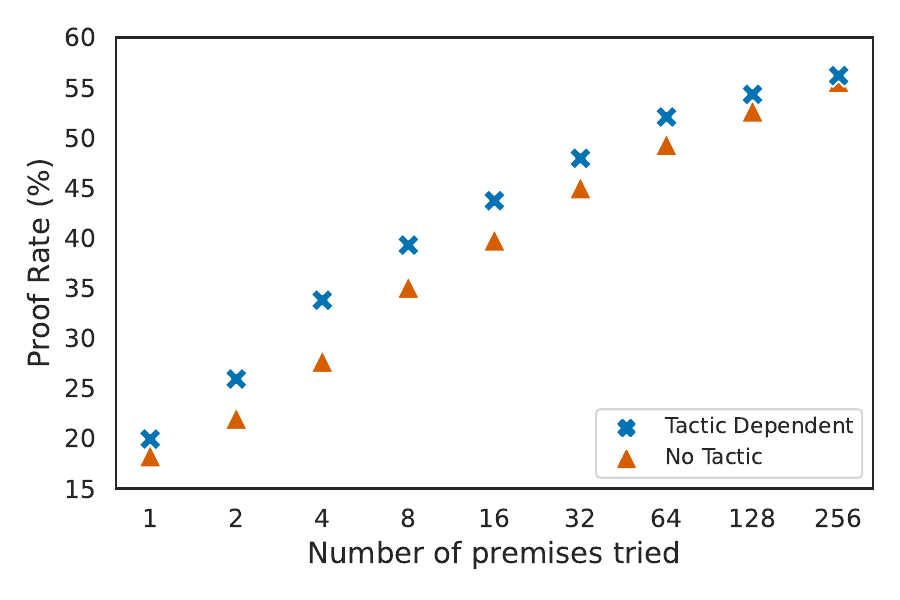}}
    \caption{We calculate \textit{accumulated proof rate} in the following way: try 1 premise, count problems solved, then try 2 premises, count problems solved using 1 or 2 premises, then try 4 premises, count problems solved using 1, 2, or 4 premises etc. Following this, on the x-axis we have the number of premises used to generate steps in Algorithm \ref{alg:evaluate_psm_on_proof_rate_benchmark}. The y-axis presents the accumulative proof rate as we try more and more premises. The higher the proof rate for the smaller number of premises used the better. We observe that prompting the model with the tactic is not necessary to achieve the final high proof. However, it allows the model for a more accurate premise selection -- all premises necessary to close the proof are ranked higher.}
    \label{fig:step_tactic_ablation}
    \end{center}
\end{figure}

We observed that different tactics use different subsets of premises. This motivated us to extend the context given to our model with \textit{tactic prompt}. Namely, provide the tactic name as an additional argument to the premise selection model, similarly to  \cite{bansal2019holist}. Prompting model with the tactic name does not yield significant improvements. However, it allows the model for a more accurate premise selection. Namely, as presented in Figure \ref{fig:step_tactic_ablation} and Table \ref{tab:num_premises_used_table_large}, we observe that premises necessary to close the proof are ranked higher. This motivates an alternative performance metric presented in the next section.


\begin{table}[H]
\caption{Effect of the number of premises used for generating tactic steps on the proof rate. We fix a set of tactics and accumulate problems solved as we increase the number of premises used to generate steps in Algorithm \ref{alg:evaluate_psm_on_proof_rate_benchmark}. Namely, for each $k$, we count the number of problems solved using at most $k$ premises.
The ``Tactic'' column indicates whether the model was given a tactic prompt.}
\label{tab:num_premises_used_table_large}
\begin{center}
\addtolength{\tabcolsep}{-3.2pt}
\begin{tabular}{lcccccccccccccc}
\toprule
Model & Tactic & Dataset & $k\leq0$ & $\leq1$ & $\leq2$ & $\leq4$ & $\leq8$ & $\leq16$ & $\leq32$ & $\leq64$ & $\leq128$ & $\leq256$ \\
\midrule

BM25 & No & N/A & 9.63 & 13.56 & 15.62 & 16.70 & 18.47 & 20.73 & 23.38 & 25.44 & 28.00 & 30.55 \\
MH-86M & No & HPL & 9.63 & 19.25 & 22.99 & 28.68 & 34.58 & 39.88 & 44.50 & 47.84 & 51.47 & 52.95 \\
MH-86M & Yes & HPL & 9.63 & 20.24 & 25.44 & 31.53 & 36.15 & 40.67 & 44.70 & 48.53 & 51.87 & 54.22 \\
MH-86M & No & MAPL & 9.63 & 18.27 & 22.00 & 27.70 & 35.07 & 39.78 & 44.99 & 49.31 & 52.65 & 55.60 \\
MH-86M & Yes & MAPL & 9.63 & 19.94 & 25.93 & 33.79 & 39.29 & 43.71 & 47.94 & 52.06 & 54.32 & 56.19 \\

\bottomrule
\end{tabular}
\addtolength{\tabcolsep}{1pt}
\end{center}
\end{table}

\subsection{Number of premises used as a performance metric for premise selection.}
\label{sec:num_premises_tried_ablation}

Consider the number of premises used to generate steps in Algorithm \ref{alg:evaluate_psm_on_proof_rate_benchmark} (parameter $k$ in the for-loop). Intuitively, the fewer premises needed the better, since it means that all the premises necessary to close the proof are ranked higher (high recall), thus the model does a more accurate premise selection. In other words, a better retrieval model should be able to score all the necessary facts higher and push unnecessary facts down the list. 

To compare different models we fix a set of tactics and accumulate problems solved as we increase the number of premises used to generate steps in Algorithm \ref{alg:evaluate_psm_on_proof_rate_benchmark}. This is presented in Table \ref{tab:num_premises_used_table_large} and Figure \ref{fig:step_tactic_ablation}.
Namely, for each $k$, we count the number of problems solved using at most $k$ premises.
Effectively, each new value of $k$ adds one new step per tactic to try.

\subsection{Single-step proof rate bound}
\label{sec:proof_rate_lower_bound}

It is non-trivial to estimate the lower bound on how many problems can be closed directly from the root state in a single proof step.
To answer this question, we use different models in Algorithm \ref{alg:evaluate_psm_on_proof_rate_benchmark} and take the union of problems solved by them. Namely, we ensemble the results of the \method variations introduced in previous sections:
\method-86M, \method-38M, \method-\methods, Sledgehammer, BM25, and the models presented in Section \ref{sec:model_size_ablation}. Such a combination successfully closes $65.5\%$ of the proofs. 


\section{Examples of proofs found by Magnushammer}

In Sledgehammer, once one of the external provers found a proof, it is likely
that it can be reproduced inside Isabelle (but not always, as reported by
\citet{Paulson2012ThreeYO}). The external provers significantly reduce the
number of premises passed to the reproduction step, therefore the Isabelle’s
proof will be short. The major bottleneck of Sledgehammer, however, is the
pre-selection step: the external provers often cannot find a proof because they
are provided too few – or too many – premises.

In Magnushammer, on the other hand, we skip the external provers completely and
input premises directly into the native Isabelle’s tactics to produce a proof.
This means that the prediction must be of high quality in order to obtain good
results. The number of the premises will be typically larger – therefore the
proofs will be longer, and of form of a combination of a strong tactic and a
long list of premises as its arguments.

As an example demonstrating the difference between Magnushammer and
Sledgehammer from the perspective of produced proofs, let’s see two proofs of
the algebraic theorem \texttt{set\_r\_ar\_cos\_ker} from the Archive of Formal
Proofs:\footnote{from the theory Group-Ring-Module/Algebra4.thy, accesible at
\url{https://search.isabelle.in.tum.de/\#theory/default\_Isabelle2022\_AFP2022/Group-Ring-Module/Algebra4}}

Sledgehammer’s proof:

\begin{verbatim}
by (smt (z3) Ring.set_r_ar_cos ker_ideal)
\end{verbatim}

Magnushammer’s proof:

\begin{verbatim}
by (clarsimp simp add: set_ar_cos_def Ring.Ring Ring.set_r_ar_cos
eq_prop Ring.I_in_set_ar_cos Set.bexE Ring.ring_is_ag ker_ideal)
\end{verbatim}

Both Sledgehammer and Magnushammer were able to solve it, however, the latter used more premises. This is expected: whenever both methods find a proof, the Magnushammer’s proof is often longer in the sense of the number of premises used. Yet, Sledgehammer’s weaker pre-selection scheme causes it to find fewer proofs in comparison.

An example of a theorem that Sledgehammer was unable to prove (with a generous
time limit of 60~s), but Magnushammer has proven, is lemma
\texttt{unit\_disc\_fix\_moebius\_uminus}.\footnote{from the theory
Complex\_Geometry/Unit\_Circle\_Preserving\_Moebius.thy, accessible at
\\\url{https://search.isabelle.in.tum.de/\#theory/default\_Isabelle2022\_AFP2022/Complex\_Geometry/Unit\_Circle\_Preserving\_Moebius}}
The proof produced by Magnushammer consists of the smt tactic and a list of
premises. Thus, Magnushammer was able to retrieve the necessary premises in
contrast to Sledgehammer:

\begin{small}
\begin{verbatim}
by (smt (z3) unit_disc_fix_unit_circle_fix
Oriented_Circlines.unit_disc_def unit_circle_fix_moebius_uminus
unit_disc_fix_moebius_comp Set.image_iff unit_disc_fix_iff
Moebius.uminus_moebius_def Unitary11_Matrices.unitary11_unitary11_gen
unit_disc_fix.abs_eq Oriented_Circlines.inf_notin_unit_disc
Moebius.plus_moebius_def unit_disc_fix_discI unit_disc_fix_moebius_add
unit_disc_fix_id_moebius Set.imageE Set.imageI
Oriented_Circlines.zero_in_unit_disc SMT.verit_minus_simplify(4)
unit_circle_fix_moebius_comp
\end{verbatim}
\end{small}


\end{document}